%% file: main.tex
\documentclass[letterpaper, 10 pt, conference]{ieeeconf}  

\IEEEoverridecommandlockouts
\usepackage{graphics}
\usepackage{epsfig}
\usepackage{mathptmx}
\usepackage{times}
\usepackage{amsmath}
\usepackage{amssymb}
\usepackage{multirow}
\usepackage{color,soul}
\usepackage{algorithm}
\usepackage{siunitx}
\usepackage{hyperref}
\usepackage{algpseudocode}
\usepackage{titlesec}
\usepackage{booktabs}
\usepackage{balance}

\usepackage[comma,numbers,compress]{natbib}

\usepackage[T1]{fontenc}
\usepackage{pifont}

\newcommand{\cmark}{\ding{51}}
\newcommand{\xmark}{\ding{55}}
\titleformat{\subsection}
  {\normalfont\bfseries} 
  {\Alph{subsection})}        
  {1em}                  
  {}                     
  
\titleformat{\subsubsection}
  {\normalfont\bfseries} 
  {\arabic{subsubsection})}        
  {1em}                  
  {}                     
  
\titleformat{\paragraph}
  {\normalfont\bfseries} 
  {\alph{paragraph})}        
  {1em}                  
  {}                     

\title{\LARGE \bf
Embedding Classical Balance Control Principles \\in Reinforcement Learning for Humanoid Recovery
}

\author{
Nehar Poddar$^{1,2}$, Stephen McCrory$^{1,2}$, Luigi Penco$^{1}$, Geoffrey Clark$^{1,2}$, Hakki Erhan Sevil$^{2}$, Robert Griffin$^{1,2}$
\thanks{$^{1}$ Florida Institute for Human and Machine Cognition, Pensacola, FL, USA.}
\thanks{$^{2}$ University of West Florida, Pensacola, FL, USA.}
\thanks{Email: \url{npoddar@ihmc.org}}
}

\begin{document}

\maketitle
\thispagestyle{empty}
\pagestyle{empty}

\input{sections/abstract}

\input{sections/introduction}

\input{sections/related_works}

\input{sections/methodology}

\input{sections/experiments_eval}

\input{sections/conclusion}

\bibliography{mybib}

\end{document}

%% file: sections/abstract.tex
\begin{abstract}
Humanoid robots remain vulnerable to falls and unrecoverable failure states, limiting their practical utility in unstructured environments. While reinforcement learning has demonstrated stand-up behaviors, existing approaches treat recovery as a pure task-reward problem without an explicit representation of the balance state. We present a unified RL policy that addresses this limitation by embedding classical balance metrics: capture point, center-of-mass state, and centroidal momentum, as privileged critic inputs and shaping rewards directly around these quantities during training, while the actor relies solely on proprioception for zero-shot hardware transfer. Without reference trajectories or scripted contacts, a single policy spans the full recovery spectrum: ankle and hip strategies for small disturbances, corrective stepping under large pushes, and compliant falling with multi-contact stand-up using the hands, elbows, and knees. Trained on the Unitree H1-2 in Isaac Lab, the policy achieves a $93.4\%$ recovery rate across randomized initial poses and unscripted fall configurations. An ablation study shows that removing the balance-informed structure causes stand-up learning to fail entirely, confirming that these metrics provide a meaningful learning signal rather than incidental structure. Sim-to-sim transfer to MuJoCo and preliminary hardware experiments further demonstrate cross-environment generalization. These results show that embedding interpretable balance structure into the learning framework substantially reduces time spent in failure states and broadens the envelope of autonomous recovery.
\end{abstract}

%% file: sections/introduction.tex
\section{Introduction}

Humanoid robots have demonstrated increasingly capable locomotion and manipulation in controlled environments \cite{radosavovic2024humanoid, haarnoja2024soccer, li2025versatile}. However, reliable recovery from falls and large disturbances remains an open challenge. Falls interrupt operation, risk hardware damage, and often require manual intervention, limiting deployment beyond laboratory settings. For humanoid platforms to operate autonomously in unstructured environments, fall recovery must be treated as a core capability rather than an exceptional case.

Fall recovery is contact-rich, non-periodic, and highly state-dependent. Depending on the disturbance, a robot must decide in real time whether to resist using ankle or hip strategies, take a stabilizing step, initiate a controlled descent, or execute a multi-contact stand-up using the hands, elbows, and knees yet no existing approach addresses all of these behaviors within a single policy. Prior works tackle isolated subproblems, including predefined stand-up sequences \cite{stuckler2006standup, huang2025host}, scripted contact schedules \cite{zhang2024wococo}, and reference motion tracking \cite{he2025gettingup}. Although effective within their respective design envelopes, these approaches require significant engineering effort and generalize poorly to unplanned contact configurations or disturbances of unexpected magnitude.

Reinforcement learning offers a path toward more general recovery \cite{gaspard2025frasa, yang2023fallrecovery, chen2025hifar}, and recent work has produced increasingly capable whole-body controllers across a range of disturbance conditions. However, most RL-based recovery formulations treat stability as an implicit consequence of the task reward, leaving the policy without an explicit representation of how close the robot is to an unrecoverable state. Without balance-aware signals, the critic must infer recoverability from reward history alone, a representation that may suffice under narrow task conditions but leaves open the question of generalization across a broader disturbance spectrum.

Classical humanoid balance analysis already provides this information in compact, interpretable form. Capture point \cite{pratt2006capture, koolen2012capturability1}, center of mass (CoM) state, and centroidal momentum together characterize whether a given configuration is recoverable under feasible contact changes and are well-established tools in model-based control \cite{pratt2012capturability2}. These quantities are readily available in simulation and can be estimated on hardware, yet they are rarely used to directly structure learned recovery policies. We argue that this represents an underexplored opportunity: embedding balance-aware structure may improve both the efficiency and generality of RL based recovery without sacrificing the adaptability of learned controllers.

\begin{figure}[t]
\centering
\includegraphics[width=1.0\linewidth]{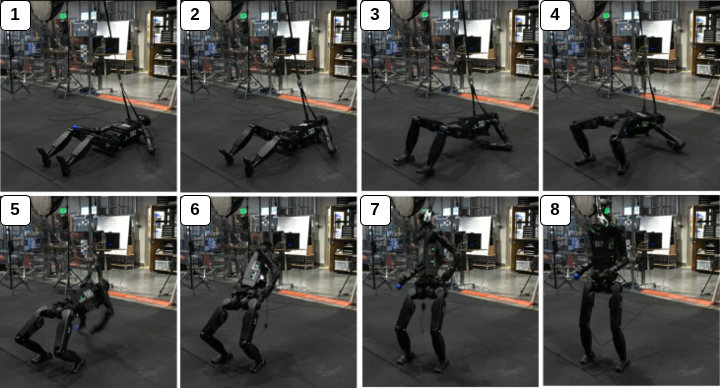}
\caption{Stand-up sequence on the Unitree H1-2 hardware. Frames 1--8 show recovery from a fallen configuration (1) to upright stance (8).}
\label{fig:grid_standup}
\end{figure}

We address this by incorporating classical balance metrics: capture point, CoM state, and centroidal momentum, directly into both the reward structure and an asymmetric critic \cite{pinto2018asymmetric}. The critic receives these quantities as privileged inputs during training \cite{kumar2021rma}, while the actor relies solely on proprioception for zero-shot hardware transfer. Without reference trajectories or scripted contacts, a single policy trained in Isaac Lab \cite{mittal2025isaac} on the Unitree H1-2 learns ankle and hip stabilization, stepping recovery, compliant falling, and multi-contact stand-up from arbitrary initial configurations.

The main contributions of this work are:
\begin{itemize}

\item A balance-informed RL formulation embedding capture point, CoM 
state, and centroidal momentum as privileged critic inputs and reward 
terms, providing explicit recoverability structure without reference 
trajectories or scripted contacts.

\item A single unified policy on the Unitree H1-2 spanning the full 
recovery spectrum: ankle and hip strategies, stepping recovery, compliant falling, and multi-contact stand-up through hands, elbows, and knees \cite{zhang2024wococo, herzog2016ankle}, trained via a curriculum that explicitly cycles fall-induction and stand-up to cover the complete 
recovery sequence.

\item An ablation study indicating that, within this formulation, 
balance-informed structure is critical for stand-up discovery: removing privileged critic inputs and capture-point rewards results in failure to leave the ground, with stuck-low termination rate rising from 0.067 to 1.0.

\item Zero-shot hardware deployment on the Unitree H1-2 across 10 trials from diverse configurations, and sim-to-sim transfer to MuJoCo, validating cross-environment generalization without policy modification.

\end{itemize}

%% file: sections/related_works.tex
\section{Related Work}
\label{sec:related-work}

\subsection{Classical Stability and Reduced-Order Models}

Early humanoid balance research established analytical criteria linking centroidal motion to contact feasibility. The Zero Moment Point (ZMP) framework provided a practical trajectory feasibility condition under fixed-foot contacts \cite{vukobratovic1972zmp}, later extended to multi-contact settings through wrench-based support region analysis \cite{bretl2008testing}. The capture point and divergent component of motion (DCM) relate CoM position and velocity to stepping requirements, unifying ankle, hip, and stepping strategies within a common framework \cite{pratt2006capture, koolen2012capturability1, herzog2016ankle, hof2005condition}. Reaction-mass and flywheel models extend these ideas to centroidal angular momentum, capturing upper-body effects \cite{lee2012momentum, gong2020angular}. Centroidal dynamics formulations then bridge reduced-order reasoning and full multi-contact whole-body behavior \cite{dai2014wholebody, sentis2010compliant}. While these methods yield interpretable stability criteria effective for structured locomotion, they rely on simplifying assumptions that become strained in the non-periodic, contact-rich scenarios characterizing post-disturbance recovery. Our work draws on these classical metrics not as controllers, but as structured signals to guide learned value estimation.

\subsection{Optimization and Multi-Contact Planning}

Optimization-based approaches formulate balance and recovery as constrained trajectory optimization problems over contact forces, centroidal dynamics, and joint motion. Whole-body trajectory optimization enables dynamically consistent multi-contact behavior \cite{dai2014wholebody}, while direct-contact formulations such as DirCon compute contact-consistent trajectories through hybrid dynamics optimization \cite{posa2016optimization}.Contact-explicit optimal control frameworks such as Crocoddyl achieve real-time feasible motion plans through efficient differential dynamic programming \cite{mastalli2020crocoddyl}. Model predictive control extends these ideas to online adaptation, with contact-implicit variants discovering contact transitions via complementarity constraints \cite{lecleach2020fast}, and reduced-order MPC schemes leveraging simplified models for push recovery and bracing \cite{yang2020bracing, kim2023mpc, zuo2024fallprotection}. Contact-Invariant Optimization treats contact activation as a decision variable to discover feasible contact phases \cite{mordatch2012discovery}, while learning-based feasibility predictors estimate whether candidate configurations are dynamically realizable \cite{carpentier2017feasibility}. Despite their expressiveness and physical grounding, these methods remain computationally demanding and challenging to apply in real time to the non-periodic, unplanned scenarios characterizing fall and recovery.

\subsection{Reinforcement Learning for Humanoid Recovery}
Reinforcement learning has been applied to fall detection, recovery control, and end-to-end stand-up using pose-tracking or task-based rewards \cite{gaspard2025frasa, yang2023fallrecovery}. Curriculum-based frameworks address high-dynamics recovery through staged training \cite{chen2025hifar}, and reward-shaping strategies have been explored to improve recovery from fallen states \cite{deng2025recover}. Several getting-up methods rely on motion references, keyframes, or canonical poses \cite{he2025gettingup, xu2025unified}, while others train reference-free, posture-spanning policies in which multi-contact strategies emerge implicitly \cite{huang2025host}. Beyond dedicated recovery controllers, recent RL methods have produced versatile humanoid behaviors including agile locomotion, running, and parkour \cite{li2025versatile, vanmarum2024reward, zhuang2024parkour}, with whole-body controllers leveraging reference motions, symmetry priors, or shared architectures to generalize across behaviors \cite{mittal2023orbit,deng2025recover, he2025hover, mittal2024symmetry}.

\begin{table}[!t]
\centering
\caption{Comparison of representative RL-based humanoid recovery approaches.}
\label{tab:standup_all}
\resizebox{\columnwidth}{!}{
\begin{tabular}{@{}lccccccc@{}}
\toprule
\textbf{Method} & \textbf{HoF} & \textbf{Fall} & \textbf{MP} 
& \textbf{MT} & \textbf{Ref} & \textbf{EB} & \textbf{AC} \\
\midrule
FRASA~\cite{gaspard2025frasa}
  & \xmark & \cmark & \xmark & \xmark & \xmark & \xmark & \xmark \\
HumanUP~\cite{he2025gettingup}
  & \cmark & \xmark & \cmark & \cmark & \cmark & \xmark & \cmark \\
HoST~\cite{huang2025host}
  & \cmark & \xmark & \cmark & \cmark & \xmark & \xmark & \xmark \\
HiFAR~\cite{chen2025hifar}
  & \cmark & \cmark & \cmark & \cmark & \xmark & \xmark & \xmark \\
\midrule
\textbf{Ours}
  & \cmark & \cmark & \cmark & \cmark & \xmark & \cmark & \cmark \\
\bottomrule
\end{tabular}
}
\vspace{3pt}
\footnotesize
\textbf{HoF}: high-DOF humanoid; 
\textbf{Fall}: fall recovery; 
\textbf{MP}: multi-pose initialization; 
\textbf{MT}: multi-terrain; 
\textbf{Ref}: uses motion reference; 
\textbf{EB}: explicit balance metrics; 
\textbf{AC}: asymmetric critic.
\vspace{-5mm}
\end{table}

In most of these formulations, balance behavior arises as a consequence of task or survival objectives rather than from explicitly structured stability criteria, as summarized in Table~\ref{tab:standup_all}. Our work is reference-free and does not rely on motion clips, keyframes, or canonical poses; instead, it incorporates CoM state, capture point, and centroidal momentum directly into the reward structure and an asymmetric critic, connecting classical stability analysis with learned recovery without predefined contact schedules.

%% file: sections/methodology.tex
\begin{figure*}[h]
    \centering
    \includegraphics[width=1.0\linewidth]{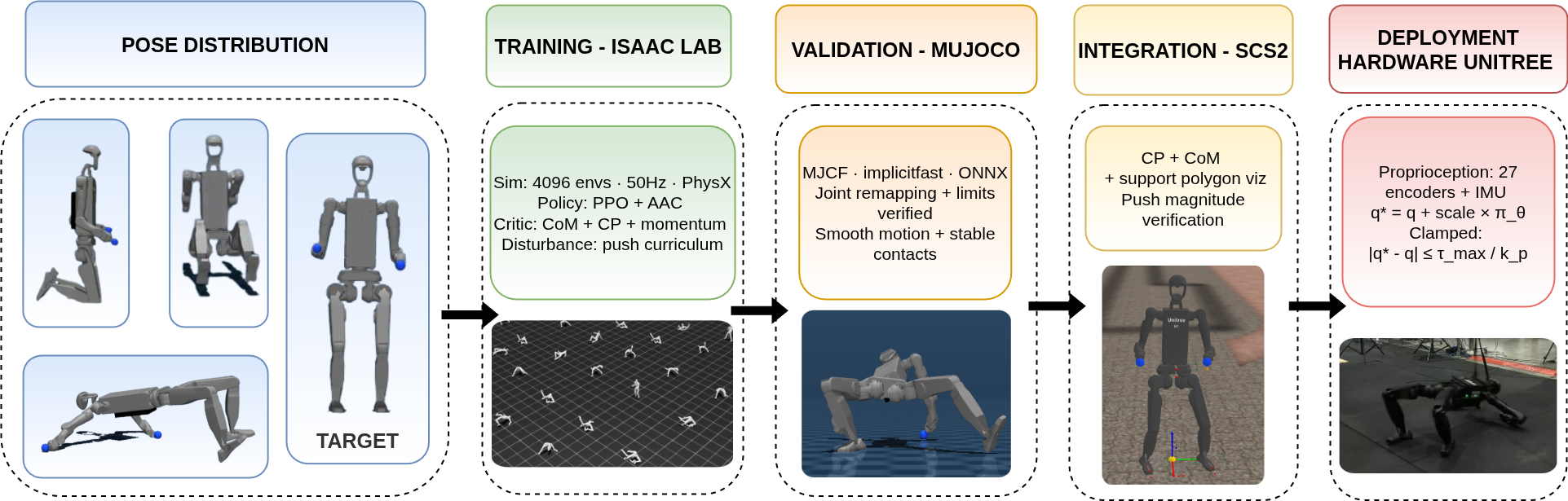}
    \caption{End-to-end pipeline for the Unitree H1-2 stand-up 
    controller: diverse initial pose distribution with upright 
    target~(left); Isaac Lab training with PPO, asymmetric 
    critic, and push curriculum; MuJoCo sim-to-sim validation; 
    SCS2 integration with balance visualization; and zero-shot 
    hardware deployment using proprioception only~(right).}
    \label{fig:pipeline}
\end{figure*}
 \section{Methodology}

\subsection{Problem Formulation}

We formulate humanoid fall recovery as a continuous-state, continuous-action Markov Decision Process (MDP) defined by $(\mathcal{S}, \mathcal{A}, P, R)$. A stochastic policy $\pi_\theta(\mathbf{a}_t \mid \mathbf{o}_t)$, parameterized by learnable weights $\theta$, maps proprioceptive observations $\mathbf{o}_t$ to joint-level actions $\mathbf{a}_t$, with the objective of maximizing the expected discounted return:
\[
J(\theta) = \mathbb{E}_{\pi_\theta} \Biggl[ \sum_{t=0}^{T} \gamma^t r_t \Biggr],
\]
where $r_t = R(\mathbf{s}_t, \mathbf{a}_t)$ is the per-step reward, $\gamma \in (0,1)$ is the discount factor, and $T$ is the episode horizon. The full system state $\mathbf{s}_t$ is available during training, while the deployed policy receives only partial observations $\mathbf{o}_t$ corresponding to onboard proprioceptive sensing. Actions are continuous joint-level position commands.

\subsection{Learning Architecture}

We employ an on-policy actor--critic framework based on Proximal Policy Optimization (PPO)~\cite{schulman2017ppo}. The actor outputs relative joint position targets for all actuated degrees of freedom of the Unitree H1-2, tracked by low-level PD controllers. The critic estimates the state-value function using privileged information available only during training, following the asymmetric actor--critic design of \cite{pinto2018asymmetric}.

Both actor and critic are multilayer perceptrons with hidden dimensions $[512, 256, 128]$ and ELU activations. The policy is initialized with action noise $\sigma = 1.0$. Training runs for 50{,}000 iterations with 24 rollout steps per environment per update. PPO hyperparameters are summarized in Table~\ref{tab:hyperparameters}. Training is conducted in Isaac Lab~\cite{mittal2023orbit}, modeling full-body dynamics, contact, joint limits, and friction.

\begin{table}[h]
\centering
\caption{PPO training hyperparameters.}
\label{tab:hyperparameters}
\begin{tabular}{lc}
\toprule
\textbf{Hyperparameter} & \textbf{Value} \\
\midrule
Iterations          & 50,000 \\
PPO clip $\epsilon$ & 0.2 \\
Discount $\gamma$   & 0.99 \\
GAE $\lambda$       & 0.95 \\
Learning rate       & $1\times10^{-3}$ \\
Target KL           & 0.01 \\
Entropy coeff.      & 0.005 \\
\bottomrule
\end{tabular}
\end{table}

\subsection{Observation Design}

To support sim-to-real transfer, we decompose the full state $\mathbf{s}_t$ into policy observations $\mathbf{o}_t$ and privileged variables available only to the critic during training.

\subsubsection{Policy Observations}
The actor receives only quantities measurable on hardware:
\begin{itemize}
    \item Base angular velocity and projected gravity,
    \item Relative joint positions and joint velocities,
    \item Previous action $\mathbf{a}_{t-1}$.
\end{itemize}
Additive uniform noise is applied to $\mathbf{o}_t$ during training to model sensor uncertainty and improve robustness to real-world measurement error.

\subsubsection{Privileged Critic Observations}
The critic receives the full actor observation $\mathbf{o}_t$ augmented with simulator-derived quantities $\mathbf{s}_t^{\text{priv}}$, forming the input $(\mathbf{o}_t, \mathbf{s}_t^{\text{priv}})$, where $\mathbf{s}_t^{\text{priv}} \subset \mathbf{s}_t$ contains quantities unavailable on hardware:
\begin{itemize}
    \item Center-of-mass (CoM) position, velocity, and acceleration,
    \item Whole-body linear and angular momentum,
    \item Capture point location.
\end{itemize}
These quantities characterize the instantaneous balance state of the robot, enabling more accurate value estimation and more stable policy updates during training.
While CoM state can in principle be estimated via IMU integration or whole-body state estimators, reliable estimation during contact-rich fall recovery is particularly challenging: drift accumulates rapidly during dynamic motion, and multi-contact estimation requires knowing which contacts to trust during uncontrolled falling.
\subsection{Action Space}

The action space comprises relative joint position commands for all actuated joints of the H1-2. At each control step, the policy outputs actions $\mathbf{a}_t$, scaled by 0.3 and executed via low-level PD controllers. To model realistic actuator and communication latencies, a randomized per-episode delay of 10--40\,ms is applied to joint commands and held constant within each episode, supporting consistent sim-to-real transfer.

\input{sections/rewards}

\subsection{Curriculum and Training Strategy}

Training follows a three-stage progressive curriculum designed to encourage broad exploration of recovery strategies, introduce physical realism, and constrain behavior to hardware-safe operating ranges.

\begin{enumerate}
    \item \textbf{Exploration phase.} Joint torque limits are initially set to $10\times$ hardware specifications, providing sufficient actuation authority for the policy to discover diverse recovery strategies, including multi-contact behaviors such as hand-assisted push-ups from the ground.

    \item \textbf{Difficulty expansion.} External disturbances, randomized initial configurations (standing, squatting, kneeling, sitting, and supine), and domain randomization over physical parameters (mass distribution, friction, inertia, and joint armature) are progressively introduced. Once the robot reaches upright stance, large destabilizing impulses are applied to induce a fall and trigger a full fall-and-recovery cycle, ensuring the policy trains on the complete recovery sequence.

    \item \textbf{Hardware constraint annealing.} Joint torque and position limits are gradually reduced toward hardware-safe values ($1\times$ torque, $0.9\times$ position limit factor), transitioning the policy from exploratory to deployable behavior.
\end{enumerate}

Episodes were terminated upon timeout or when unsafe 
torso--ground contact occurred under high contact force 
conditions, encouraging recovery while discouraging 
catastrophic collapse.

\subsection{Domain Randomization and Noise Injection}

To improve sim-to-real robustness, domain randomization is applied across dynamics, contacts, initialization, observations, and disturbances:

\begin{itemize}
    \item \textbf{Actuation and joint dynamics:} Stiffness and damping gains scaled by $[0.75, 1.25]$; joint friction sampled from $[0.5, 4.0]$; armature scaled by $[0.5, 1.6]$.

    \item \textbf{Contact properties:} Static friction $\mu_s \sim [0.3, 1.6]$, dynamic friction $\mu_d \sim [0.3, 1.2]$, restitution $e \sim [0, 0.2]$.

    \item \textbf{Initialization noise:} Base pose perturbations $\Delta p \in [-0.05, 0.05]$\,m, $\Delta r \in [-0.2, 0.2]$\,rad; joint offsets $\Delta q \in [-0.1, 0.1]$\,rad.

    \item \textbf{Observation noise:} Additive uniform noise on policy inputs: base angular velocity $\omega_b \pm 0.5$\,rad/s, projected gravity $\pm 0.15$, joint positions $q \pm 0.1$\,rad, joint velocities $\dot{q} \pm 0.5$\,rad/s.

    \item \textbf{External disturbances:} Random torso impulses of 50--300\,N over 0.1--0.2\,s for push recovery training. Once upright stance is reached, large destabilizing impulses are applied to induce a fall and trigger a full fall-and-recovery cycle.
\end{itemize}

Rigid-body inertias are randomized without recomputing a dynamically consistent model, requiring the policy to infer balance-relevant structure directly from observed state trajectories rather than relying on accurate internal dynamics.

%% file: sections/rewards.tex
\subsection{Physics-Guided Reward Design}
\label{sec:reward}

We decompose the total reward into three physically motivated 
groups:
\begin{equation}
    R(\mathbf{s}_t)
    = R^{\mathrm{I}}(\mathbf{s}_t)
    + R^{\mathrm{II}}(\mathbf{s}_t)
    + R^{\mathrm{III}}(\mathbf{s}_t)
    \label{eq:total_reward}
\end{equation}
where $R^{\mathrm{I}}$ enforces vertical recovery, 
$R^{\mathrm{II}}$ enforces balance and capturability, and 
$R^{\mathrm{III}}$ provides regularization and motion priors. 
This grouping mirrors the physical recovery sequence: vertical 
elevation $\rightarrow$ dynamic balance $\rightarrow$ motion 
refinement. Table~\ref{tab:notation} summarizes all symbols 
and Table~\ref{tab:reward} provides a full decomposition.

\begin{table}[t]
\centering
\renewcommand{\arraystretch}{1.15}
\setlength{\tabcolsep}{4.5pt}
\caption{Summary of symbols used in the reward formulation.}
\label{tab:notation}
\resizebox{\columnwidth}{!}{
\begin{tabular}{@{}lll@{}}
\toprule
\textbf{Category} & \textbf{Symbol} & \textbf{Description} \\
\midrule
\multirow{2}{*}{State}
 & $\mathbf{s}_t$   & Robot state at discrete time $t$ \\
 & $\Delta t$       & Control/simulation time step \\
\midrule
\multirow{6}{*}{CoM}
 & $\mathbf{p}_c \in \mathbb{R}^3$ & 3-D CoM position \\
 & $\mathbf{p}_{c,xy} \in \mathbb{R}^2$ & Horizontal CoM projection \\
 & $h_c$     & Vertical CoM height ($p_{c,z}$) \\
 & $h^{*}$   & Target upright CoM height \\
 & $\dot{h}_c$ & Vertical CoM velocity \\
 & $e_h$     & Height error $(h_c - h^{*})$ \\
\midrule
\multirow{3}{*}{Balance}
 & $\boldsymbol{\xi}$ & Capture point (LIP model) \\
 & $g$               & Gravitational acceleration \\
 & $\mathbb{I}[\cdot]$ & Indicator function \\
\midrule
\multirow{4}{*}{Support}
 & $\mathcal{C}$ & Convex hull of active contacts \\
 & $\mathcal{C}_{\mathrm{feet}}$ & Convex hull of foot contacts \\
 & $d_{\mathrm{CoM}}$ & Signed distance: CoM projection to $\mathcal{C}$ \\
 & $d_{\mathrm{CP}}$  & Signed distance: CP to $\mathcal{C}_{\mathrm{feet}}$ \\
\midrule
\multirow{4}{*}{Momentum}
 & $\mathbf{p}_{\ell}$ & Whole-body linear momentum \\
 & $\mathbf{L}$        & Whole-body angular momentum \\
 & $\mathbf{F}_{\mathrm{net}}$ & Net centroidal force 
    ($\dot{\mathbf{p}}_{\ell}/\Delta t$) \\
 & $\boldsymbol{\tau}_{\mathrm{net}}$ & Net centroidal torque 
    ($\dot{\mathbf{L}}/\Delta t$) \\
\midrule
\multirow{4}{*}{Parameters}
 & $w_{\bullet}$      & Group weighting coefficients \\
 & $\alpha_{\bullet}$ & Shaping coefficients \\
 & $\sigma_{\bullet}$ & Gaussian bandwidth parameters \\
 & $\delta_h$         & Height deadband threshold \\
\bottomrule
\end{tabular}
}
\vspace{-3mm}
\end{table}

\subsubsection{Group I: Vertical Recovery}

Group~I rewards drive vertical recovery toward a target CoM 
height $h^*$:
\begin{equation}
    R^{\mathrm{I}}(\mathbf{s}_t)
    = w_h\, r_{\mathrm{height}}
    + w_r\, r_{\mathrm{rise}}
    + w_f\, r_{\mathrm{fall}}
    + w_s\, r_{\mathrm{stab}}
    \label{eq:group1}
\end{equation}

\paragraph{Height tracking.}
A dense Gaussian reward centered at $h^*$ provides a smooth 
gradient throughout the elevation range:
\begin{equation}
    r_{\mathrm{height}}
    = \exp\!\left(-\frac{(h_c - h^{*})^2}{\sigma_h^2}\right)
    \label{eq:rheight}
\end{equation}
where $\sigma_h$ controls the reward bandwidth.

\paragraph{Vertical momentum shaping.}
$r_{\mathrm{rise}}$ rewards upward CoM velocity when the robot 
is below target height; $r_{\mathrm{fall}}$ quadratically 
penalizes downward velocity to discourage collapse:
\begin{align}
    r_{\mathrm{rise}}
    &= \alpha_r\,\max(\dot{h}_c,\,0)\,
       \mathbb{I}[h_c < h^{*}]
    \label{eq:rrise} \\
    r_{\mathrm{fall}}
    &= -\alpha_f\,\max(-\dot{h}_c,\,0)^2
    \label{eq:rfall}
\end{align}
This asymmetric structure rewards rising proportionally to 
upward velocity while penalizing downward motion quadratically. 
Neither term activates near $h^*$, avoiding conflicting 
incentives during final convergence.

\paragraph{Stationary stability near target.}
Once the robot approaches $h^*$, $r_{\mathrm{stab}}$ provides 
a proximity bonus within tolerance $\delta_h$ while penalizing 
residual vertical velocity to suppress overshoot:
\begin{equation}
    r_{\mathrm{stab}}
    = \alpha_s\,\mathbb{I}\!\left[|h_c - h^{*}| < \delta_h\right]
    - \alpha_v\,|\dot{h}_c|
    \label{eq:rstab}
\end{equation}

The Gaussian in~\eqref{eq:rheight} provides a nonzero gradient 
at all heights below $h^*$, and $r_{\mathrm{rise}}$ encourages 
upward momentum throughout, giving the policy a continuous 
learning signal across the full range of initial CoM heights 
rather than only at discrete canonical poses.

\subsubsection{Group II: Balance and Capturability}

Group~II rewards enforce static and dynamic stability by 
embedding capture-point theory directly into the reward 
structure:
\begin{equation}
    R^{\mathrm{II}}(\mathbf{s}_t)
    = w_c\, r_{\mathrm{CoM}}
    + w_{\xi}\, r_{\mathrm{CP}}
    + w_m\, r_{\mathrm{mom}}
    \label{eq:group2}
\end{equation}

Capture Point: Balance is assessed using the capture point 
$\boldsymbol{\xi}$, derived from the Linear Inverted 
Pendulum (LIP) model~\cite{pratt2006capture}, which 
indicates the location where the robot must step to 
arrest a fall:
\begin{equation}
    \boldsymbol{\xi}
    = \mathbf{p}_{c,xy}
    + \frac{\dot{\mathbf{p}}_{c,xy}}{\sqrt{g/h_c}}
    \label{eq:cp}
\end{equation}
The signed distances $d_{\mathrm{CoM}}$ and $d_{\mathrm{CP}}$ 
measure how far the CoM projection and capture point lie from 
the boundaries of $\mathcal{C}$ and 
$\mathcal{C}_{\mathrm{feet}}$ respectively (negative inside, 
positive outside).

\paragraph{Static Stability}
$r_{\mathrm{CoM}}$ rewards configurations in which the CoM 
projection lies within the support hull $\mathcal{C}$:
\begin{equation}
    r_{\mathrm{CoM}}
    = \exp\!\left(-\frac{d_{\mathrm{CoM}}^2}{\sigma_c^2}\right)
    \label{eq:rcom}
\end{equation}
The smooth Gaussian decay encourages stable base maintenance 
without hard threshold switching.

\paragraph{Dynamic Capturability}
$r_{\mathrm{CP}}$ penalizes configurations in which the capture 
point falls outside the foot support hull 
$\mathcal{C}_{\mathrm{feet}}$, indicating that balance cannot 
be maintained under the LIP model without a contact 
change~\cite{koolen2012capturability1}:
\begin{equation}
    r_{\mathrm{CP}}
    = \exp\!\left(-\frac{d_{\mathrm{CP}}^2}{\sigma_{\xi}^2}\right)
    \label{eq:rcp}
\end{equation}
Together, $r_{\mathrm{CoM}}$ and $r_{\mathrm{CP}}$ couple 
static and dynamic stability: the former penalizes 
instantaneous support violations while the latter penalizes 
momentum states that may lead to future balance loss even when 
the CoM projection currently lies within the support hull.

\paragraph{Momentum Regularization}
$r_{\mathrm{mom}}$ penalizes large net centroidal forces and 
moments to suppress impulsive whole-body reactions:
\begin{equation}
    r_{\mathrm{mom}}
    = -\alpha_{\ell}\,\|\mathbf{F}_{\mathrm{net}}\|_2^2
    - \alpha_{L}\,\|\boldsymbol{\tau}_{\mathrm{net}}\|_2^2
    \label{eq:rmom}
\end{equation}
Quadratic penalization disproportionately discourages large 
impulsive reactions relative to small smooth corrections.

\subsubsection{Group III: Regularization and Motion Shaping}

Group~III introduces hardware safety constraints and 
physically consistent motion priors:
\begin{equation}
    R^{\mathrm{III}}(\mathbf{s}_t)
    = r_{\mathrm{torque}}
    + r_{\mathrm{orient}}
    + r_{\mathrm{contact}}
    + r_{\mathrm{alive}}
    \label{eq:group3}
\end{equation}

Safety terms penalize torque and joint limit violations and 
rapid action changes, protecting hardware during high-force 
recovery transitions. The posture term penalizes torso 
misalignment with gravity, encouraging upright orientation 
throughout recovery. The contact shaping term applies 
height-gated rewards for hand and elbow support, promoting 
appropriate multi-contact transitions during stand-up. 
The survival term combines an alive bonus with an early 
termination penalty, encouraging task completion rather 
than episode length maximization.

Taken together, Group~III terms encourage the detailed 
recovery sequence: including multi-contact transitions 
and compliant falling, to emerge from the physics-based 
objectives in $R^{\mathrm{I}}$ and $R^{\mathrm{II}}$, 
without scripted contact schedules or reference trajectories.

\begin{table}[t]
\centering
\caption{Physics-guided reward decomposition.}
\label{tab:reward}
\renewcommand{\arraystretch}{1.15}
\setlength{\tabcolsep}{5pt}
\begin{tabular}{@{}llll@{}}
\toprule
\textbf{Group} & \textbf{Component} & \textbf{Symbol} 
& \textbf{Objective} \\
\midrule
\multirow{4}{*}{$R^{\mathrm{I}}$}
 & Height tracking & $r_{\mathrm{height}}$ & Reach $h^{*}$ \\
 & Rise shaping    & $r_{\mathrm{rise}}$   & Encourage upward motion \\
 & Fall penalty    & $r_{\mathrm{fall}}$   & Suppress downward velocity \\
 & Stability band  & $r_{\mathrm{stab}}$   & Prevent oscillation \\
\midrule
\multirow{3}{*}{$R^{\mathrm{II}}$}
 & CoM support     & $r_{\mathrm{CoM}}$    & Static stability \\
 & Capture point   & $r_{\mathrm{CP}}$     & Dynamic capturability \\
 & Momentum reg.   & $r_{\mathrm{mom}}$    & Smooth centroidal dynamics \\
\midrule
\multirow{4}{*}{$R^{\mathrm{III}}$}
 & Safety          & $r_{\mathrm{torque}}$  & Hardware protection \\
 & Posture         & $r_{\mathrm{orient}}$  & Upright torso alignment \\
 & Contact shaping & $r_{\mathrm{contact}}$ & Multi-contact transitions \\
 & Survival        & $r_{\mathrm{alive}}$   & Task completion \\
\bottomrule
\end{tabular}
\vspace{-5mm}
\end{table}

%% file: sections/experiments_eval.tex
\section{Experiments and Results}

\subsection{Simulation}

All policies were trained and evaluated in Isaac Lab using 
the Unitree H1-2 model, with additional validation in MuJoCo 
to assess cross-simulator generalization. Under nominal 
conditions, the policy maintained stable standing for the 
full episode horizon in both environments.

\subsubsection{Recovery from Diverse Configurations} 
The policy successfully recovered to upright stance from a variety of initial poses, including squatting, kneeling, sitting, and supine, without relying on predefined trajectories. Across 10,000 trials, it achieved a 93.4\% recovery success rate with a mean recovery time of 5\,s. Failures primarily occurred when the robot became trapped in local contact configurations from which upright recovery could not be completed within the episode horizon.

\subsubsection{Disturbance Rejection}
External perturbations of increasing magnitude were applied 
during standing. Small pushes (50--100\,N) were absorbed 
through ankle strategies; moderate pushes (100--200\,N) 
induced hip-based adjustments; and larger pushes 
(200--300\,N) triggered stepping responses. This graduated 
hierarchy is consistent with the balance structure embedded 
in the reward and with classical push-recovery 
analysis~\cite{herzog2016ankle}. For disturbances 
$\geq$300\,N, the robot transitioned to multi-contact 
recovery, bracing with the hands before returning to 
upright stance.
\begin{figure}[t]
    \centering
    \includegraphics[width=\columnwidth]{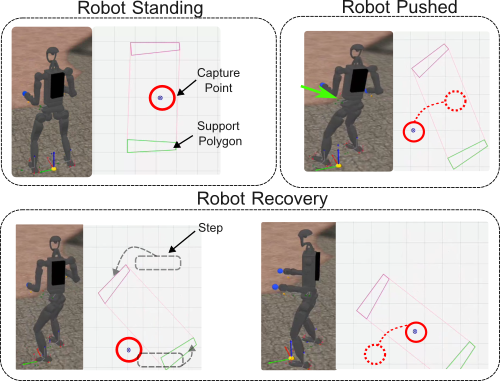}
    \caption{Capture-point evolution during push recovery. 
    The blue marker denotes $\boldsymbol{\xi}$ and polygons 
    represent active foot support regions. The policy 
    progressively drives $\boldsymbol{\xi}$ back inside the 
    support hull through ankle, hip, and stepping responses.}
    \label{fig:capture_point_recovery}
    \vspace{-5mm}
\end{figure}
Fig.~\ref{fig:capture_point_recovery} shows the capture 
point evolution during a representative push recovery trial. 
The policy drives $\boldsymbol{\xi}$ back inside the foot 
support hull through the ankle, hip, and stepping sequence, 
consistent with the capturability reward structure in 
Eqs.~\eqref{eq:rcom}--\eqref{eq:rcp}.

\subsubsection{Sim-to-Sim Validation}
To assess cross-simulator generalization, we evaluated the
policy in MuJoCo across five initial configurations
(standing, squatting, kneeling, supine, and sitting) under
applied forces from 0--500\,N in four lateral directions.
MuJoCo's more conservative contact model, with stricter
friction and damping defaults than Isaac Lab, we observe, more closely
approximates physical hardware dynamics, providing a
stronger test of transferability than in-distribution
Isaac Lab evaluation alone.

A trial is counted as successful only if the CoM exceeds
0.85\,m and remains there for at least 1\,s with vertical
velocity below 0.1\,m/s and torso tilt within 72$^\circ$
of vertical, ruling out momentary bounces or transient
uprighting. Without perturbation, the policy achieves
100\% recovery from all poses. Under perturbation,
robustness degrades gracefully: standing and squatting
maintain recovery rates above 50\% at 200\,N, while
supine and sitting drop below 25\% beyond 100\,N,
consistent with the greater centroidal momentum that must
be regulated from near-floor orientations. Recovery time
increases with force magnitude, reaching up to 9\,s for
supine trials at high forces.

\begin{figure*}[t]
    \centering
    \includegraphics[width=1.0\linewidth]{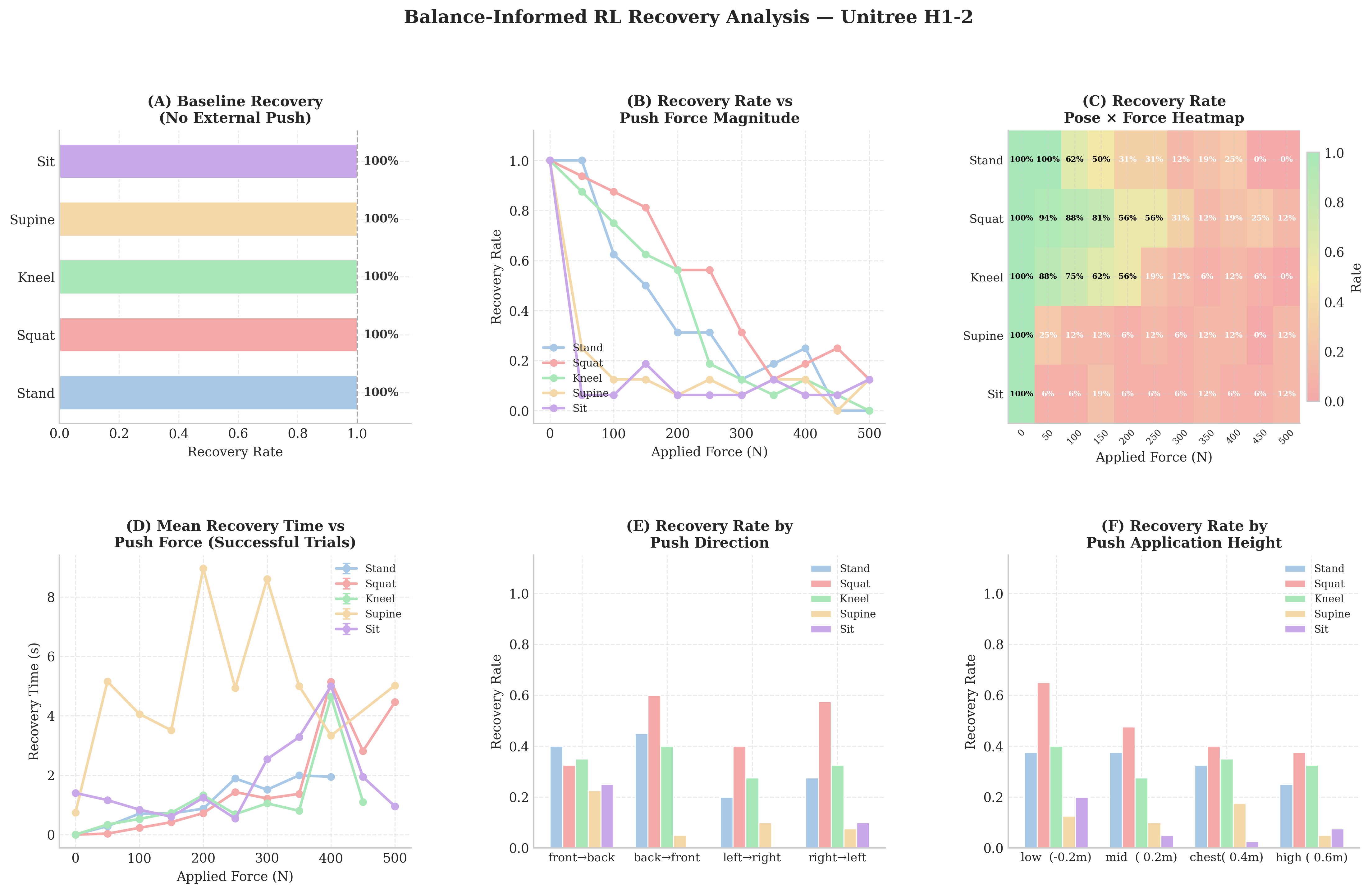}
    \caption{MuJoCo sim-to-sim recovery analysis across 
    five initial poses and forces 0--500\,N. Success 
    requires the CoM to exceed 0.85\,m and remain stable 
    for $\geq$1\,s. (A) Baseline recovery without push. 
    (B--C) Recovery rate degrades gracefully with force; 
    standing and squatting are most robust. (D) Recovery 
    time increases with perturbation magnitude. (E--F) 
    Directional and height sensitivity show consistent 
    trends across poses.}
    \label{fig:recovery_analysis}
    \vspace{-5mm}
\end{figure*}

Push application height has a consistent effect: low pushes
($-$0.2\,m) yield the highest recovery rates while high
pushes (0.6\,m) yield the lowest. A force applied above
the CoM induces larger whole-body angular momentum, driving
greater displacement of $\boldsymbol{\xi}$ outside the
support polygon, the precise failure mode penalised by
Eq.~\eqref{eq:rcp}. Results are summarized in
Fig.~\ref{fig:recovery_analysis}.

\subsection{Ablation Study}

To evaluate the contribution of our key design choices, we 
trained a reduced policy without privileged critic inputs 
(CoM dynamics, capture point, centroidal momentum), 
capture-point reward terms, or progressive curriculum, with 
all other hyperparameters and reward terms held identical.

Results are shown in Table~\ref{tab:ablation}. The reduced 
policy failed to learn stand-up behavior: episodes 
terminated exclusively via the stuck-low condition 
(triggered when CoM height shows no increase for 
2 consecutive seconds) at a rate of 1.0, compared to 
0.067 in the full policy, indicating the robot never left 
the ground. Mean reward fell from $+$379.2 to $-$115.3 
and all recovery metrics collapsed. These results suggest 
that, within this training setup, balance-informed structure 
is necessary for the policy to discover multi-contact 
stand-up behavior. We note that this is a single combined 
ablation; disentangling the individual contributions of 
privileged critic inputs, capture-point rewards, and 
curriculum is left for future work.

\begin{table}[t]
\centering
\caption{Ablation at convergence: full policy vs.\ reduced 
variant (no privileged critic, no capture-point rewards, 
no curriculum).}
\label{tab:ablation}
\renewcommand{\arraystretch}{1.12}
\setlength{\tabcolsep}{5pt}
\begin{tabular}{@{}lcc@{}}
\toprule
\textbf{Metric} & \textbf{Full} & \textbf{Reduced} \\
\midrule
\multicolumn{3}{@{}l}{\textbf{Performance}} \\
Mean reward             & $\mathbf{+379.2}$ & $-115.3$ \\
Episode length (steps)  & $\mathbf{975}$    & $219$ \\
\multicolumn{3}{@{}l}{\textbf{Recovery}} \\
CoM height maintenance  & $\mathbf{13.23}$  & $0.17$ \\
Progressive recovery    & $\mathbf{3.78}$   & $0.41$ \\
Hand-support recovery   & $\mathbf{1.80}$   & $0.11$ \\
Feet under CoM          & $\mathbf{1.58}$   & $0.00$ \\
\multicolumn{3}{@{}l}{\textbf{Termination}} \\
Timeout                 & $\mathbf{0.934}$  & $0.000$ \\
Stuck low               & $0.067$           & $\mathbf{1.000}$ \\
\bottomrule
\end{tabular}
\vspace{-5mm}
\end{table}

\subsection{Hardware Deployment}

Policies trained in simulation were exported as ONNX 
networks and deployed on the physical Unitree H1-2 using 
our software stack, integrated via the OpenCV DNN 
Java API at 50\,Hz. The joint-level control law was
\[
\tau = k_p\left(q_{\mathrm{ref}} - q\right) - k_d\dot{q},
\]
where $q_{\mathrm{ref}}$ is the policy output and $q$, 
$\dot{q}$ are measured joint position and velocity. 
No modifications were made to the policy or control 
structure at deployment.

Across 10 trials initialized from supine, seated, 
kneeling, and crouched configurations, the robot 
successfully achieved upright recovery in all 10 attempts 
with no parameter tuning between simulation and hardware. 
In 3 trials, transient high-frequency joint oscillations were observed following recovery, prompting conservative safety intervention. We attribute this behavior to residual sim-to-real mismatch in damping characteristics and are addressing it through improved action-smoothing during training.

During disturbance trials, the robot successfully rejected 
manual pushes and recovered balance. Observable increases 
in ankle and hip motor torques following perturbations 
were consistent with the ankle and hip stabilization 
strategies learned in simulation, providing qualitative 
evidence that the balance hierarchy transfers to hardware.

%% file: sections/conclusion.tex
\section{Conclusion and Future Work}

We presented a balance-informed reinforcement learning framework for humanoid fall recovery that incorporates classical stability metrics: capture point, CoM state, and whole-body momentum, as privileged critic inputs and reward terms. Rather than enforcing predefined contact sequences, this structure gives the critic an explicit representation of recoverability, enabling the actor to decide when to resist a disturbance, when to step, and when to engage multi-contact support.

Simulation results demonstrate a $93.4\%$ recovery rate across diverse initial configurations, while an ablation without balance-informed structure fails entirely to learn stand-up behavior. These results suggest that the embedded metrics provide a meaningful learning signal rather than an incidental cue. A single policy spans the full recovery spectrum: ankle and hip strategies, stepping recovery, compliant falling, and hand-assisted stand-up, all achieved without reference trajectories or scripted contact schedules. Zero-shot hardware deployment on the Unitree H1-2 successfully recovers from diverse configurations across 10 trials, providing initial validation that the learned balance structure transfers to real hardware.

These findings highlight a broader insight: classical balance metrics offer a compact, transferable representation for contact-rich humanoid control that complements purely task-driven reward design. Whether this structure generalizes beyond fall recovery: to manipulation under external loads or locomotion over complex terrain, remains an open question. In particular, maintaining capturability and regulating momentum may serve as prerequisites for robust manipulation, allowing a robot to respond to externally applied loads without explicit payload estimation, potentially simplifying downstream control.

Future work will explore perception-driven contact selection in cluttered environments, multi-surface recovery, integration with long-horizon locomotion and manipulation tasks, and tighter coupling between balance-aware control and task-level planning. Extending to non-planar surfaces is a natural next step; however, the convex hull support polygon metric assumes coplanar contacts, and replacing it with a more general criterion, such as wrench-space feasibility over non-coplanar contact sets, would be necessary. By using a planar segmentation representation of the environment, the policy could infer contact normals and select contacts within a continuous action space.

%% file: mybib.bib
@inproceedings{gaspard2025frasa,
  title={FRASA: An End-to-End Reinforcement Learning Agent for Fall Recovery and Stand Up of Humanoid Robots},
  author={Clément Gaspard and Marc Duclusaud and Grégoire Passault and Mélodie Daniel and Olivier Ly},
  year={2025},
booktitle={IEEE International Conference on Robotics and Automation (ICRA)},
}

@inproceedings{vanmarum2024reward,
  title={Revisiting Reward Design and Evaluation for Robust 
         Humanoid Standing and Walking},
  author={van Marum, Bart and Shrestha, Aayam and Duan, Helei 
          and Dugar, Pranay and Dao, Jeremy and Fern, Alan},
  booktitle={IEEE/RSJ International Conference on Intelligent 
             Robots and Systems (IROS)},
  year={2024}
}

@inproceedings{mittal2024symmetry,
  title={Symmetry Considerations for Learning Task Symmetric Robot Policies},
  author={Mayank Mittal and Nikita Rudin and Victor Klemm and Arthur Allshire and Marco Hutter},
  year={2024},
booktitle={IEEE International Conference on Robotics and Automation (ICRA)},

}

@inproceedings{chen2025hifar,
  title={HiFAR: Multi-Stage Curriculum Learning for High-Dynamics Humanoid Fall Recovery},
  author={Penghui Chen and Yushi Wang and Changsheng Luo and Wenhan Cai and Mingguo Zhao},
  year={2025},
booktitle={IEEE/RSJ International Conference on Intelligent Robots and Systems (IROS)},
}

@inproceedings{he2025hover,
  title={Hover: Versatile neural whole-body controller for humanoid robots},
  author={He, Tairan and Xiao, Wenli and Lin, Toru and Luo, Zhengyi and Xu, Zhenjia and Jiang, Zhenyu and Kautz, Jan and Liu, Changliu and Shi, Guanya and Wang, Xiaolong and others},
  booktitle={IEEE International Conference on Robotics and Automation (ICRA)},
  year={2025},
  organization={IEEE}
}

@inproceedings{he2025gettingup,
  title={Learning Getting-Up Policies for Real-World Humanoid Robots},
  author={Xialin He and Runpei Dong and Zixuan Chen and Saurabh Gupta},
  year={2025},
  booktitle={Robotics: Science and Systems (RSS)},
}

@article{haarnoja2024soccer,
  title={Learning Agile Soccer Skills for a Bipedal Robot with Deep Reinforcement Learning},
  author={Haarnoja, Tuomas and Moran, Ben and Lever, Guy and others},
  journal={Science Robotics},
  volume={9},
  number={89},
  year={2024}
}

@misc{zhuang2024parkour,
  title={Humanoid Parkour Learning},
  author={Ziwen Zhuang and Shenzhe Yao and Hang Zhao},
  year={2024},
  eprint={2406.10759},
  archivePrefix={arXiv},
  primaryClass={cs.RO}
}

@article{mittal2023orbit,
  author={Mittal, Mayank and Yu, Calvin and Yu, Qinxi and Liu, Jingzhou and Rudin, Nikita and Hoeller, David and Yuan, Jia Lin and Singh, Ritvik and Guo, Yunrong and Mazhar, Hammad and Mandlekar, Ajay and Babich, Buck and State, Gavriel and Hutter, Marco and Garg, Animesh},
  journal={IEEE Robotics and Automation Letters},
  title={Orbit: A Unified Simulation Framework for Interactive Robot Learning Environments},
  year={2023},
  volume={8},
  number={6},
}

@article{zuo2024fallprotection,
  author={Zuo, Weilong and Gao, Junyao and Liu, Jiongnan and Wu, Taiping and Xin, Xilong},
  title={Whole-Body Dynamics for Humanoid Robot Fall Protection Trajectory Generation with Wall Support},
  journal={Biomimetics},
  volume={9},
  number={4},
  year={2024},
}

@article{sentis2010compliant,
  author={Luis Sentis and Jaeheung Park and Oussama Khatib},
  title={Compliant Control of Multicontact and Center-of-Mass Behaviors in Humanoid Robots},
  journal={IEEE Transactions on Robotics},
  volume={26},
  number={3},
  year={2010},
}

@article{deng2025recover,
  title={Learning to Recover: Dynamic Reward Shaping with Wheel-Leg Coordination for Fallen Robots},
  author={Boyuan Deng and Luca Rossini and Jin Wang and Weijie Wang and Nikolaos Tsagarakis},
  year={2025},
    journal={arXiv preprint arXiv:2506.05516},

}

@inproceedings{pratt2006capture,
  title={Capture Point: A Step Toward Humanoid Push Recovery},
  author={Pratt, Jerry and Carff, John and Drakunov, Sergey and Goswami, Ambarish},
  booktitle={IEEE-RAS International Conference on Humanoid Robots},
  year={2006}
}

@article{koolen2012capturability1,
  title={Capturability-based Analysis and Control of Legged Locomotion, Part 1: Theory and Application to Three Simple Gait Models},
  author={Koolen, Twan and de Boer, Tomas and Rebula, John and Goswami, Ambarish and Pratt, Jerry},
  journal={International Journal of Robotics Research},
  volume={31},
  number={9},
  year={2012}
}

@article{pratt2012capturability2,
  title={Capturability-based Analysis and Control of Legged Locomotion, Part 2: Application to M2V2, a Lower-Body Humanoid},
  author={Pratt, Jerry and Koolen, Twan and de Boer, Tomas and Rebula, John and Cotton, Sebastien and Carff, John and Johnson, Matthew and Neuhaus, Peter},
  journal={International Journal of Robotics Research},
  volume={31},
  number={10},
  year={2012}
}

@article{lee2012momentum,
  title={A momentum-based balance controller for humanoid robots on non-level and non-stationary ground},
  author={Lee, Sung-Hee and Goswami, Ambarish},
  journal={Autonomous Robots},
  volume={33},
  number={4},
  year={2012}
}

@article{gong2020angular,
  title={Angular Momentum About the Contact Point for Control of Bipedal Locomotion: Validation in a LIP-based Controller},
  author={Gong, Yukai and Grizzle, Jessy},
  journal={arXiv preprint arXiv:2008.10763},
  year={2020}
}

@article{vukobratovic1972zmp,
  title={Contribution to the synthesis of biped gait},
  author={Vukobratovic, M. and Stepanenko, B.},
  journal={RAIRO Automatique},
  year={1972}
}

@article{yang2020bracing,
  title={Bracing for Impact: Robust Humanoid Push Recovery with Reduced Order Models},
  author={Yang, L. and Werner, B. and Ghansah, A.B. and Ames, A.D.},
  journal={IEEE Transactions on Robotics},
  year={2020}
}

@inproceedings{lecleach2020fast,
  title={Fast Contact-Implicit Model-Predictive Control},
  author={Le Cleac'h, S. and others},
  booktitle={IEEE Robotics and Automation Letters},
  year={2020}
}

@article{bretl2008testing,
  title={Testing static equilibrium for legged robots},
  author={Bretl, Timothy and Lall, Sanjay},
  journal={IEEE Transactions on Robotics},
  year={2008}
}

@inproceedings{huang2025host,
  title={Learning Humanoid Standing-up Control across Diverse Postures},
  author={Huang, Tao and Ren, Junli and Wang, Huayi and Wang, Zirui and Ben, Qingwei and Wen, Muning and Chen, Xiao and Li, Jianan and Pang, Jiangmiao},
  booktitle={Robotics: Science and Systems (RSS)},
  year={2025}
}

@inproceedings{dai2014wholebody,
  title={Whole-body motion planning with centroidal dynamics and full kinematics},
  author={Dai, Hongkai and Valenzuela, Andrés and Tedrake, Russ},
  booktitle={IEEE-RAS International Conference on Humanoid Robots},
  year={2014},
}

@article{kim2023mpc,
  title={A Model Predictive Capture Point Control Framework for Robust Humanoid Balancing via Ankle, Hip, and Stepping Strategies},
  author={Kim, Myeong-Ju and Lim, Daegyu and Park, Gyeongjae and Lee, Kwanwoo and Park, Jaeheung},
  journal={IEEE Transactions on Robotics},
  year={2025},
  publisher={IEEE}
  }

@article{radosavovic2024humanoid,
  title={Real-World Humanoid Locomotion with Reinforcement Learning},
  author={Radosavovic, Ilija and Xiao, Tete and Zhang, Bike and Darrell, Trevor and Malik, Jitendra and Sreenath, Koushil},
  journal={Science Robotics},
  volume={9},
  year={2024}}

@inproceedings{stuckler2006standup,
  title={Getting Back on Two Feet: Reliable Standing-Up Routines for a Humanoid Robot},
  author={St{\"u}ckler, J{\"o}rg and Schwenk, Johannes and Behnke, Sven},
  booktitle={International Conference on Intelligent Autonomous Systems},
  year={2006}
}

@article{li2025versatile,
  title={Reinforcement Learning for Versatile, Dynamic, and Robust Bipedal Locomotion Control},
  author={Li, Zhongyu and Peng, Xue Bin and Abbeel, Pieter and Levine, Sergey and Berseth, Glen and Sreenath, Koushil},
  journal={International Journal of Robotics Research},
  volume={44},
  year={2025}
}

@inproceedings{zhang2024wococo,
  title={{WoCoCo}: Learning Whole-Body Humanoid Control with Sequential Contacts},
  author={Zhang, Chong and Xiao, Wenli and He, Tairan and Shi, Guanya},
  booktitle={Conference on Robot Learning (CoRL)},
  year={2024}
}

@article{yang2023fallrecovery,
  title={Learning Complex Motor Skills for Legged Robot Fall Recovery},
  author={Yang, Chengxu and Pu, Chuanyu and Xin, Guiyang and Zhang, Jie and Li, Zhibin},
  journal={IEEE Robotics and Automation Letters},
  volume={8},
  number={7},
  year={2023},
}

@inproceedings{pinto2018asymmetric,
  title={Asymmetric Actor Critic for Image-Based Robot Learning},
  author={Pinto, Lerrel and Andrychowicz, Marcin and Welinder, Peter and Zaremba, Wojciech and Abbeel, Pieter},
  booktitle={Robotics: Science and Systems (RSS)},
  year={2018}
}

@inproceedings{kumar2021rma,
  title={{RMA}: Rapid Motor Adaptation for Legged Robots},
  author={Kumar, Ashish and Fu, Zipeng and Pathak, Deepak and Malik, Jitendra},
  booktitle={Robotics: Science and Systems (RSS)},
  year={2021}
}

@inproceedings{herzog2016ankle,
  title={Momentum Control with Hierarchical Inverse Dynamics on a Torque-Controlled Humanoid},
  author={Herzog, Alexander and Rotella, Nicholas and Mason, Sean and Grimminger, Felix and Schaal, Stefan and Righetti, Ludovic},
  booktitle={IEEE-RAS International Conference on Humanoid Robots},
  year={2016}
}

@article{hof2005condition,
  title={Condition for Dynamic Stability},
  author={Hof, At L.},
  journal={Journal of Biomechanics},
  volume={38},
  number={1},
  year={2005},
}

@inproceedings{posa2016optimization,
  title={Optimization and Stabilization of Trajectories for Constrained Dynamical Systems},
  author={Posa, Michael and Tobenkin, Mark and Tedrake, Russ},
  booktitle={IEEE International Conference on Robotics and Automation (ICRA)},
  year={2016},
}

@inproceedings{mastalli2020crocoddyl,
  title={Crocoddyl: An Efficient and Versatile Framework for Multi-Contact Optimal Control},
  author={Mastalli, Carlos and Budhiraja, Rohan and Merkt, Wolfgang and Saurel, Guilhem and Hammoud, Bilal and Naveau, Maximilien and Carpentier, Justin and Righetti, Ludovic and Vijayakumar, Sethu and Mansard, Nicolas},
  booktitle={IEEE International Conference on Robotics and Automation (ICRA)},
  year={2020},
}

@article{mordatch2012discovery,
  title={Discovery of Complex Behaviors through Contact-Invariant Optimization},
  author={Mordatch, Igor and Todorov, Emanuel and Popovic, Zoran},
  journal={ACM Transactions on Graphics},
  volume={31},
  number={4},
  year={2012},
}

@inproceedings{carpentier2017feasibility,
  title={Learning Feasibility Constraints for Multicontact Locomotion of Legged Robots},
  author={Carpentier, Justin and Budhiraja, Rohan and Mansard, Nicolas},
  booktitle={Robotics: Science and Systems (RSS)},
  year={2017},
}

@article{xu2025unified,
  title={Unified Humanoid Fall-Safety Policy from a Few Demonstrations},
  author={Xu, Zhengjie and Li, Ye and Lin, Kwan-yee and Yu, Stella X.},
  journal={arXiv preprint arXiv:2511.07407},
  year={2025}
}

@article{schulman2017ppo,
  title={Proximal Policy Optimization Algorithms},
  author={Schulman, John and Wolski, Filip and Dhariwal, 
          Prafulla and Radford, Alec and Klimov, Oleg},
  journal={arXiv preprint arXiv:1707.06347},
  year={2017}
}

@article{mittal2025isaac,
  title={Isaac lab: A gpu-accelerated simulation framework for multi-modal robot learning},
  author={Mittal, Mayank and Roth, Pascal and Tigue, James and Richard, Antoine and Zhang, Octi and Du, Peter and Serrano-Munoz, Antonio and Yao, Xinjie and Zurbr{\"u}gg, Ren{\'e} and Rudin, Nikita and others},
  journal={arXiv preprint arXiv:2511.04831},
  year={2025}
}
